\documentclass[a4paper,twocolumn,10pt]{article}
\usepackage[top=2cm,bottom=2.1cm,left=1.8cm,right=1.8cm]{geometry}
\usepackage[T1]{fontenc}
\usepackage{lmodern}
\usepackage[english]{babel}
\usepackage{amsmath,amssymb,mathtools}
\usepackage{graphicx,booktabs,tabularx,array}
\usepackage{enumitem}
\usepackage{xcolor}
\usepackage{tikz}
\usetikzlibrary{arrows.meta,positioning,calc,fit,backgrounds}
\usepackage[colorlinks=true,linkcolor=blue!45!black,citecolor=blue!45!black,urlcolor=blue!45!black]{hyperref}
\usepackage[font=small,labelfont=bf,labelsep=period]{caption}
\usepackage{titlesec}
\titleformat{\section}{\large\bfseries}{\thesection}{0.6em}{}
\titleformat{\subsection}{\normalsize\bfseries}{\thesubsection}{0.6em}{}
\titleformat{\paragraph}[runin]{\normalsize\bfseries}{}{0pt}{}[.]
\titlespacing*{\section}{0pt}{1.0\baselineskip}{0.45\baselineskip}
\titlespacing*{\subsection}{0pt}{0.75\baselineskip}{0.35\baselineskip}
\titlespacing*{\paragraph}{0pt}{0.5\baselineskip}{0.5em}
\setlist{nosep,leftmargin=*,topsep=0.25\baselineskip}
\newcolumntype{Y}{>{\raggedright\arraybackslash}X}
\definecolor{raaNavy}{HTML}{193D57}
\definecolor{raaTeal}{HTML}{277A78}
\definecolor{raaGold}{HTML}{A66C2D}
\definecolor{raaInk}{HTML}{243847}
\definecolor{raaPale}{HTML}{EFF4F7}
\definecolor{raaMint}{HTML}{EDF6F3}
\definecolor{raaSand}{HTML}{FAF3E9}

\title{Autonomy and Coordination in Intelligent Manufacturing: \\ A Recursive Autonomous Agent Framework}
\author{%
Yuning Chen\textsuperscript{1,\textdagger}, Hongda Zhang\textsuperscript{1}, Linqiang Hu\textsuperscript{1}, Ziqing Zhou\textsuperscript{1}, Yi Liu\textsuperscript{1},\\
Zhao Huang\textsuperscript{2}, Ding Zhou\textsuperscript{2}, Haining Xu\textsuperscript{2,*}\\[0.5em]
{\small\textsuperscript{1}Fudan University\quad
\textsuperscript{2}Jiangsu Chuangling Intelligent Technology Co., Ltd.}%
}
\date{}

\begin{document}
\makeatletter
\twocolumn[{%
\begin{@twocolumnfalse}
\maketitle
\begin{abstract}
Intelligent production requires systems to make decisions and organize work according to production goals and operating conditions. It also requires continuous adjustment through feedback. This paper analyzes intelligence in human production organizations from a complex adaptive systems (CAS) perspective. The analysis identifies two coupled mechanisms: individual autonomy and collective coordination. Individual autonomy enables actors to form local decisions using available information. Collective coordination aligns these decisions with shared goals and joint constraints. Learning and adaptation improve both mechanisms over time. A production-system model formalizes how local decision rules and coordination jointly form system decisions. On this basis, we propose the Recursive Autonomous Agent (RAA) framework, which combines goal-driven autonomous agents with a recursive organizational structure. Goal contracts define local objectives and operating boundaries, while capability contracts support the selection and composition of production units. Common interfaces allow coordinated groups to participate as units in larger organizations, connecting equipment, production lines, and factories. RAA thus provides an engineering approach to next-generation intelligent manufacturing, translating the requirements of intelligence, openness, and system-wide integration into mechanisms for autonomous and coordinated production.
\end{abstract}
\noindent\textbf{Keywords:} intelligent manufacturing; complex adaptive systems; autonomous agent; Recursive Autonomous Agent
\vspace{1em}
\end{@twocolumnfalse}
}]
\makeatother
\begingroup
\renewcommand{\thefootnote}{\fnsymbol{footnote}}
\footnotetext[2]{First author: Yuning Chen. \href{mailto:chenyuning@fudan.edu.cn}{chenyuning@fudan.edu.cn}.}
\footnotetext[1]{Corresponding author: Haining Xu. \href{mailto:xuhaining@jscltec.com}{xuhaining@jscltec.com}.}
\endgroup

\section{Introduction}\label{sec:intro}

A factory must turn changing orders into products that meet delivery, quality, and safety requirements. It must also deal with changes in equipment and resources. A fixed program can handle repeated operations under familiar conditions. A new order, a failed machine, or a material delay may require a different plan. People then have to assess the situation and organize a response. This need becomes more pressing as product variety grows and production batches become smaller~\cite{mousavi2023,nie2023}.
Digital systems make production easier to describe and monitor. Networks make information available across equipment and departments. Digital twins strengthen the link between physical operations and their digital representations~\cite{soori2023}. These capabilities support decisions, but access to information does not itself decide how production should proceed. The practical question is how a system can use the information it has to choose actions and revise them as conditions change.

The question has both a local and a collective side. A workstation needs to understand its task, select suitable operations, and check the result. A production line must also manage shared resources and dependencies between workstations. A locally sensible choice can disrupt another task. The ability of each unit to act and the ability of the organization to coordinate those actions must therefore be developed together.
Human production organizations offer a useful reference. An operator observes conditions at a station and adjusts work within an assigned responsibility. A team leader handles issues that affect several stations. A manager connects these arrangements to broader production objectives. Problems and results move upward, while goals and work assignments move downward. This organization combines independent judgment with coordinated action. It provides a functional reference for engineering, rather than a requirement to reproduce human cognition.

We examine this reference through complex adaptive systems (CAS) theory~\cite{holland2006,levin1998}. CAS focuses on interacting actors that adapt using local information and feedback. It also draws attention to outcomes that depend on how actors interact. This perspective connects the decisions made by production units to the performance of the organization as a whole.
Our central account has two parts. \emph{Individual autonomy} explains how production actors form and revise local decisions. \emph{Collective coordination} explains how these decisions become joint action under shared constraints. Learning and adaptation develop both mechanisms. On this basis, we present the Recursive Autonomous Agent (RAA) framework. It combines goal-driven agents with a recursive structure for organizing their work.

The paper makes two contributions. First, it develops a CAS-based account of production intelligence and expresses the roles of autonomy and coordination in a production-system model. Second, it maps these roles to engineering requirements and a recursive agent framework. The design links local execution to broader production goals through contracts, feedback, and coordination across levels.

Section~\ref{sec:mechanisms} explains the two mechanisms. Section~\ref{sec:model} gives their mathematical description. Section~\ref{sec:design} develops the engineering framework and its workflow. Section~\ref{sec:manufacturing} relates the design to manufacturing paradigms and their practical requirements. Section~\ref{sec:conclusion} concludes the paper.

\section{CAS Analysis of Production Intelligence}\label{sec:mechanisms}

This section examines intelligence in human production organizations from a complex adaptive systems (CAS) perspective. The analysis identifies two coupled mechanisms: individual autonomy and collective coordination. Learning and adaptation improve both mechanisms over time. The following subsections explain why these mechanisms are needed and specify the functional requirements for their engineering implementation.

\subsection{Individual Autonomy}

Production information is distributed across different roles and organizational levels. Operators know the current condition of workpieces and equipment at their stations. Team leaders track progress across their teams. Factory managers oversee production as a whole, but cannot monitor every local detail in real time. Observations can also be delayed or inaccurate. A decision process that waits for a complete global picture may respond too slowly to changes on the shop floor.

Within the production organization considered here, actors use the information available where they work. They need enough authority to make local decisions and correct local problems. Their actions remain subject to assigned goals and operating boundaries. An issue outside those boundaries must be passed to a level that can address it.

A skilled worker illustrates the required cycle. The worker checks the equipment and workpiece, decides how to perform the operation, carries it out, and inspects the result. Feedback informs the next action. If a deviation can be corrected locally, the worker makes the correction. Otherwise, the issue is reported for coordination. The engineering requirement is a similar connection between observation, decision, execution, and feedback.

\begin{quote}
\textbf{Individual autonomy} is the ability of a production actor to use local information to pursue an assigned goal within defined boundaries. The actor maintains a local feedback loop and handles execution deviations within its authority.
\end{quote}

Four requirements follow from this definition:
\begin{description}[style=unboxed,leftmargin=0pt]
\item[AR1: Goal understanding.] The actor receives and interprets its assigned objective.
\item[AR2: Local closed loop.] Observation, decision, execution, and feedback remain connected during operation.
\item[AR3: Deviation handling.] The actor corrects local execution problems within its authority.
\item[AR4: Controlled learning.] The actor improves its decisions through experience without bypassing operating boundaries.
\end{description}

These requirements concern more than the ability to issue an action. They connect the assigned goal to the observed result. They also distinguish a correction made during execution from a change to the model or policy used in later work.

\subsection{Collective Coordination}

Autonomous action alone does not organize a production system. Two units may need the same machine at the same time. A downstream station may depend on an upstream task that has not finished. A handoff may fail even when both stations can perform their own operations. These dependencies make the relationship between actions part of the decision problem.

There are two reasons to coordinate. The first is joint feasibility. An action can be allowed at each station in isolation but disallowed when the stations act together. The second is the dependence of overall results on interaction. Having every machine available does not guarantee timely production. Task order, resource allocation, and handoffs can change the outcome of the whole process.

\begin{quote}
\textbf{Collective coordination} is the ability of several autonomous actors to form jointly feasible system decisions in pursuit of a shared goal. It establishes the relationships needed for joint work and adjusts or ends them when conditions change.
\end{quote}

This mechanism has four corresponding requirements:
\begin{description}[style=unboxed,leftmargin=0pt]
\item[CR1: Shared goals.] The organization identifies a common objective and explains how local tasks contribute to it.
\item[CR2: Capability matching.] It brings together actors with suitable capabilities and specifies how they will work together.
\item[CR3: Joint decisions.] It coordinates local proposals into feasible system action and supports their execution.
\item[CR4: Adaptive coordination.] It uses feedback to assess collective results and revise the organization of work.
\end{description}

Coordination therefore includes both preparing an organization and managing its operation. Suitable participants must first be found. Their proposed actions must then satisfy shared constraints. Execution feedback may require changes to task assignments, resources, or coordination relationships.

\begin{figure*}[t]
    \centering
    \includegraphics[width=\textwidth]{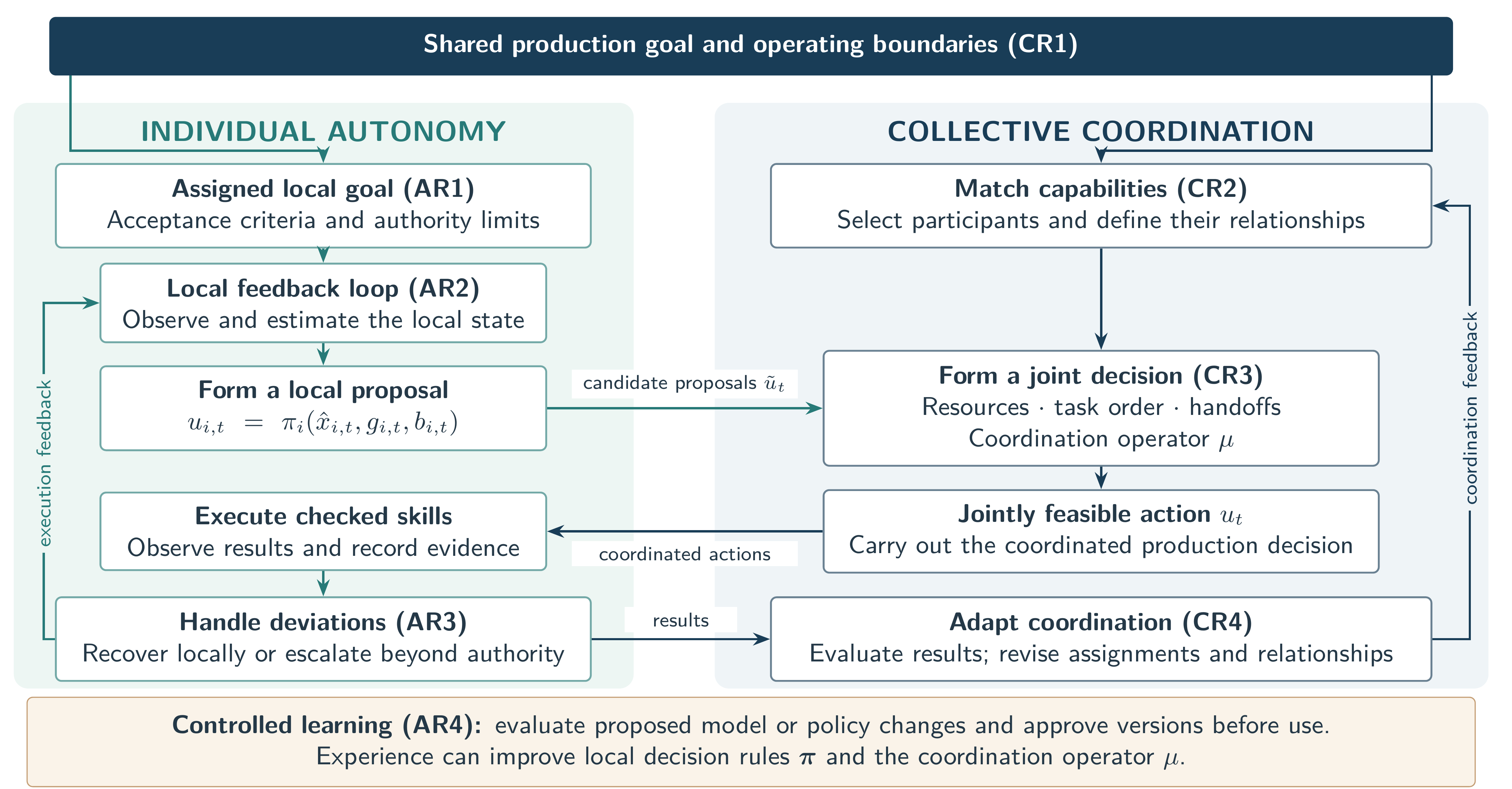}
    \caption{Two connected decision processes. Each actor uses the local process shown on the left. Individual autonomy forms local proposals through a feedback loop. Collective coordination checks their combination against shared goals and joint constraints. Execution results support local recovery, controlled learning, and changes to coordination.}
    \label{fig:mechanisms}
\end{figure*}

\subsection{Coupled Mechanisms}

The two mechanisms answer different questions. Individual autonomy asks what an actor should do using the information it has. Collective coordination asks how the proposed actions can work together. Figure~\ref{fig:mechanisms} shows the connection. Local decisions become proposals for joint work. Shared constraints determine which combinations can be executed. The resulting feedback informs both the actors and their organization.

Without individual autonomy, a central system would have to guide each local response and could become slow when conditions change. Without collective coordination, actors could compete for resources or pursue incompatible local objectives. A complete account of production decisions must cover both the actors' internal activity and their relationships.

At this level of description, the paper treats autonomy and coordination as a minimal and complete pair. The first covers the internal path from a local goal to an outcome. The second covers the relationships needed for a joint outcome. Sensing, planning, execution, and learning belong to the actor's activity. Task allocation, negotiation, shared resources, and feedback aggregation concern relationships among actors. 
Learning and adaptation operate within both parts of this account. An actor can improve a local model or decision rule. An organization can improve participant selection or change a coordination relationship. These processes develop autonomy and coordination over time; they are not introduced here as a third mechanism alongside them.

\section{Production Decision Model}\label{sec:model}

The preceding account can be expressed in a common model. The model identifies the production state, the decisions available to the system, and the connection between local and collective decisions. It describes the roles that an implementation must realize. 

\subsection{System Formulation}

Let a production system be
\begin{equation}
\mathcal P=\langle\mathcal A,\mathcal T,\mathcal R,\mathcal E,
\mathcal X,\mathcal U,\mathcal D,\mathcal G,\mathcal B,\Phi\rangle.
\label{eq:system}
\end{equation}
Here, $\mathcal A$ denotes production actors, $\mathcal T$ denotes tasks, and $\mathcal R$ denotes resources. The set $\mathcal E$ describes relationships such as assignments and process dependencies. The remaining terms describe operation: $\mathcal X$ is the state space, $\mathcal U$ is the decision space, and $\mathcal D$ is the disturbance space. Goals belong to $\mathcal G$, and operating boundaries belong to $\mathcal B$. The function $\Phi$ describes state transitions.

At discrete time $t$, write the state as
\begin{equation}
x_t=\langle x_t^{\mathrm A},x_t^{\mathrm T},x_t^{\mathrm R},x_t^{\mathrm E}\rangle\in\mathcal X.
\label{eq:state}
\end{equation}
The four components record actor capabilities and availability, task progress, resource conditions, and current relationships. Resource conditions include location, sufficiency, and occupancy. Relationships include task assignments, process dependencies, and cooperation between actors. Together, these components describe the production situation at the current time.

The system faces a goal $g_t\in\mathcal G$ and boundaries $b_t\in\mathcal B$. It selects a decision $u_t\in\mathcal U$. A disturbance $d_t\in\mathcal D$, such as a machine failure or a material delay, also affects what happens next:
\begin{equation}
x_{t+1}=\Phi(x_t,u_t,d_t;g_t,b_t).
\label{eq:transition}
\end{equation}
At this decision point, $x_t$ has already arisen and the incoming disturbance is not chosen by the system. Goals and boundaries can be revised at a higher level, but they are given within the current cycle. Under these conditions, $u_t$ is the quantity the system can actively select to influence its next state.

The system acts on an estimate $\hat x_t$, which may be incomplete or inaccurate. Its decision rule is
\begin{equation}
\begin{aligned}
u_t&=\Gamma(\hat x_t,g_t,b_t),\\
u_t&\in\mathcal U^{\mathrm{adm}}(x_t,b_t).
\end{aligned}
\label{eq:rule}
\end{equation}
Here, $\Gamma$ denotes the system-level decision rule. It maps the estimated state $\hat x_t$, the production goal $g_t$, and the operating boundaries $b_t$ to the decision $u_t$.
The admissible set $\mathcal U^{\mathrm{adm}}$ contains decisions allowed by the true state and current boundaries. Equation~\eqref{eq:rule} states two requirements: form a goal-directed decision from available information, and ensure that the chosen action respects the applicable boundaries. It specifies what a decision process must achieve. The equation alone does not provide a way to guarantee admissibility under imperfect information.

In this model, production intelligence acts through decision formation and revision. Data, models, and communication support intelligence when they help choose an appropriate $u_t$ and use execution feedback to improve subsequent choices.

\subsection{Local Decisions and Joint Constraints}

Actor $a_i$ forms a proposed action using its local state estimate, objective, and boundaries:
\begin{equation}
u_{i,t}=\pi_i(\hat x_{i,t},g_{i,t},b_{i,t}),\qquad i=1,\ldots,n.
\label{eq:local}
\end{equation}
The rule $\pi_i$ represents local decision formation. The collection of proposals gives a candidate joint decision,
\begin{equation}
\tilde u_t=(u_{1,t},u_{2,t},\ldots,u_{n,t}).
\label{eq:candidate}
\end{equation}
These proposals are the local components from which the system decision is formed. Their direct combination is not necessarily executable.

Let $\mathcal U_i(x_{i,t})$ contain actions feasible for actor $a_i$ in its local state. The Cartesian product of these sets contains all combinations of locally feasible actions. If shared constraints exclude some combinations, then the jointly feasible set satisfies
\begin{equation}
\mathcal U^{\mathrm J}(x_t)\subsetneq\prod_{i=1}^{n}\mathcal U_i(x_{i,t}).
\label{eq:joint}
\end{equation}
For example, each of two stations may be able to use one resource, while simultaneous use is forbidden. Task precedence and spatial conflicts create similar restrictions. The final system action must account for these constraints rather than simply adopting $\tilde u_t$.

Overall outcomes also depend on the combination of actions. If $y_{i,t}$ is a local outcome and $y_t$ is the system outcome, in general $y_t\neq\sum_i y_{i,t}$. Separate improvements at each station do not necessarily yield a corresponding production-line improvement. A poor handoff or a resource conflict may offset them. This dependence on interaction is the production meaning of emergence used in the paper.

\subsection{System Decision Rule}

Collect the local rules in $\boldsymbol\pi$ and use $\mu$ to represent coordination:
\begin{equation}
\begin{aligned}
\boldsymbol\pi&=(\pi_1,\pi_2,\ldots,\pi_n),\\
\mathfrak I_{\mathrm P}&=\langle\boldsymbol\pi,\mu\rangle,\qquad
\Gamma=\mu\circ\boldsymbol\pi.
\end{aligned}
\label{eq:combined}
\end{equation}
The pair $\mathfrak I_{\mathrm P}$ expresses the two mechanisms. Local rules produce candidate actions; coordination determines how they form a system decision under shared constraints. The composition $\Gamma$ is the rule in Equation~\eqref{eq:rule}. It connects local judgment to collective action.

Learning can change the local rules in $\boldsymbol\pi$. Feedback about joint outcomes can also change $\mu$, for example by informing task allocation or coordination protocols. This gives a common description of improvement at the individual and organizational levels.

\section{RAA Framework}\label{sec:design}

The model separates two engineering tasks: building units that can act autonomously, and organizing those units into a coordinated system. RAA addresses both through agents that expose compatible interfaces at different levels.

\begin{quote}
\textbf{The Recursive Autonomous Agent (RAA) framework} organizes intelligent production through goal-driven agents operating within explicit constraints. Individual agents provide local autonomy, while recursive organizations provide collective coordination.
\end{quote}

An agent associated with equipment is called a level-one agent. A level-two agent organizes several level-one agents. The level-two agent and its members can then appear as one unit to a higher level. The pattern connects equipment to production lines and production lines to factories. It also preserves a common way to state goals, describe capabilities, and report results.

Table~\ref{tab:requirements} connects the requirements in Section~\ref{sec:mechanisms} to the engineering design. These are functional requirements. Different models, planners, and communication systems may be used to implement them.

\begin{table*}[t]
\caption{From decision requirements to engineering mechanisms.}
\label{tab:requirements}
\centering
\small
\renewcommand{\arraystretch}{1.15}
\begin{tabularx}{\textwidth}{@{}p{1.1cm}p{3.05cm}Y Y@{}}
\toprule
Code & Requirement & Engineering mechanism & Role in production\\
\midrule
AR1 & Goal understanding & Goal contract with acceptance criteria and boundaries & Define the result to achieve and the allowed scope of action\\
AR2 & Local closed loop & Sensing, local model, planning, checked skill execution, and feedback & Connect local conditions to action and outcome checks\\
AR3 & Deviation handling & Recovery, replanning, safe stopping, and escalation & Resolve local problems or refer them to the appropriate level\\
AR4 & Controlled learning & Evaluated and approved model or policy updates & Improve capability while preserving operating boundaries\\
\midrule
CR1 & Shared goals & Goal decomposition and subtask acceptance & Link local work to the overall production objective\\
CR2 & Capability matching & Capability contracts and participant selection & Compose suitable units without binding tasks to fixed devices\\
CR3 & Joint decisions & Shared-constraint checks and coordination protocols & Manage resource use, task order, and handoffs\\
CR4 & Adaptive coordination & Outcome evaluation and organizational adjustment & Revise assignments, resources, or relationships as conditions change\\
\bottomrule
\end{tabularx}
\end{table*}

\subsection{Autonomous Agent Design}

\paragraph{Start with a goal contract}
A conventional automated device receives an action command and follows a program. In RAA, an agent receives a goal and the conditions under which it may act. The goal contract states the desired result and its acceptance criteria. It also specifies timing, priority, and the safety, quality, and authority limits that apply. This makes goal understanding operational: the agent has a result to achieve and explicit boundaries for its choices.

\paragraph{Connect observation to verified execution}
Multimodal sensing captures equipment and environmental conditions. A local world model interprets this information. Planning and scheduling turn the goal into a sequence of skills and decide when to use them. The execution module invokes those skills through equipment interfaces. Results and execution evidence return to the agent, which compares them with the intended outcome before selecting its next action.

The design separates artificial intelligence (AI) proposals from physical execution. AI can interpret a task and propose a plan. Before an action reaches equipment, it must pass capability and physical-feasibility checks. The framework does not allow probabilistic model output to drive equipment without those checks. This separation connects flexible planning to the deterministic requirements of execution.

\paragraph{Recover within authority}
A constraint engine monitors operation. When results depart from the plan, the agent first checks whether it can recover within its authority. It may retry an operation or revise its plan. Where permitted, it can use an alternative method or continue with reduced capability. If it cannot resolve the problem within its limits, it escalates the issue or stops safely. The event and the actions taken are recorded.

Recovery has a local scope. The agent does not resolve a cross-unit problem by exceeding its authority. Instead, it supplies the information needed by the coordinating level. The distinction allows quick local responses while keeping broader decisions with the organization responsible for them.

\paragraph{Control changes to learned capability}
Operational correction and learning run on different time scales. The fast process uses existing models, skills, and boundaries to correct deviations or replan. The slower process collects experience and proposes changes to a model, rule, or policy. Proposed updates pass offline evaluation, simulation, constraint checks, and version approval before use. Experience can therefore improve later decisions without allowing the update process to bypass physical operating limits.

\begin{figure*}[t]
    \centering
    \includegraphics[width=\textwidth]{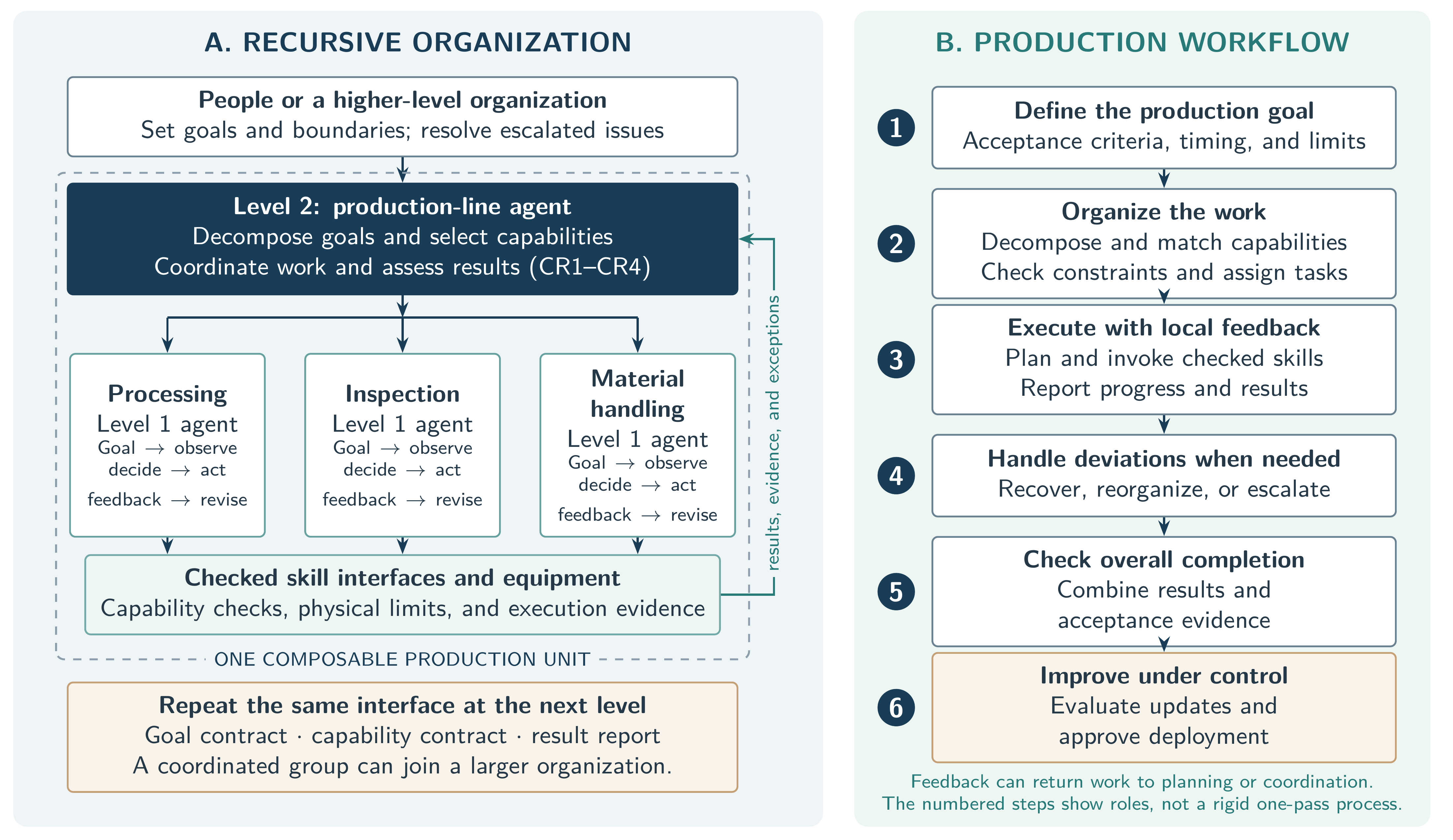}
    \caption{RAA organization and operation in a functional production-line example. Left: a level-two agent coordinates processing, inspection, and material-handling agents through compatible goal, capability, and result interfaces. Together they can be used as one unit at a higher level. Right: the execution sequence, including feedback, escalation, and controlled improvement.}
    \label{fig:workflow}
\end{figure*}

\subsection{Recursive Organization}

A level-two agent receives a production objective and divides it into interdependent subtasks. Each subtask has an acceptance criterion, so its contribution to the overall result can be checked. Completion is assessed through the decomposition and the associated requirements, rather than from an informal report that each participant has finished.

The organization finds participants through capability contracts. Each contract states what an agent can do, the conditions it needs, the effects and quality it can provide, and the resources it uses. The level-two agent combines this information with current availability and state. It then selects participants, assigns roles, and establishes working relationships.

The planning question is expressed in terms of required capabilities. A task does not have to remain tied to a particular device. A replacement device can participate when its interfaces are adapted and its capabilities meet the relevant contract. Capability verification remains necessary; a matching name alone is not enough to establish that two devices can perform the same work.

Level-one agents propose local actions. The level-two agent checks their combination against resource restrictions, task order, and handoff requirements. It forms an executable joint decision and coordinates the resulting work. Protocols specify how participants exchange state, hand over tasks, and confirm results. These mechanisms implement the role of $\mu$ in Equation~\eqref{eq:combined}.

During execution, participants report their states, results, exceptions, and supporting evidence. The level-two agent uses these reports in its own observation--decision--execution--feedback loop. It assesses progress and overall outcomes. It can revise assignments or resource allocation, change participants, or adjust coordination relationships. Completed coordination records provide experience for later selection and organization decisions.

Recursion follows from the external interface. Both a level-one agent and a level-two organization can receive goals and constraints, describe capabilities, and report results. A higher-level agent can therefore call on a coordinated group as one production unit. At each level, goals are decomposed downward and results are brought together upward. The same organizational pattern can be repeated at a broader production scale.

\subsection{Operational Workflow}

Figure~\ref{fig:workflow} illustrates the design with three equipment tasks: processing, inspection, and material handling. Each level-one agent maintains its local feedback loop. A level-two agent coordinates their work toward a production-line objective. Personnel or an existing management system set the overall goal and constraints.

The processing agent observes the workpiece and tool, chooses an operation, and checks quality and completion. The inspection agent compares measurements with acceptance criteria and returns pass or fail evidence. The material-handling agent checks locations and path conditions. It selects a route and handoff, then confirms delivery. Each agent invokes validated skills within physical operating limits.

The operation has six stages:
\begin{enumerate}
\item \textbf{Specify the outcome.} Personnel or the management system supply the production goal and its boundaries.
\item \textbf{Organize the work.} The level-two agent decomposes the goal, matches subtasks to capabilities, checks joint constraints, and assigns work.
\item \textbf{Execute locally.} Each level-one agent acts within its boundaries and reports state, progress, and evidence of the result.
\item \textbf{Handle exceptions at the right level.} Local deviations are corrected locally. Issues affecting coordination go to the level-two agent. Matters beyond that agent's authority go to personnel.
\item \textbf{Check completion.} The level-two agent brings together subtask results and acceptance evidence. It checks the overall production requirements and reports the outcome.
\item \textbf{Improve under control.} Execution and coordination records inform evaluated updates to individual agents and the organization.
\end{enumerate}

This arrangement connects management decisions to physical action through feedback at each level. People set goals and boundaries, resolve escalated issues, and make high-risk decisions. They need not continuously direct every permitted local adjustment.

\begin{figure*}[t]
    \centering
    \includegraphics[width=0.7\textwidth]{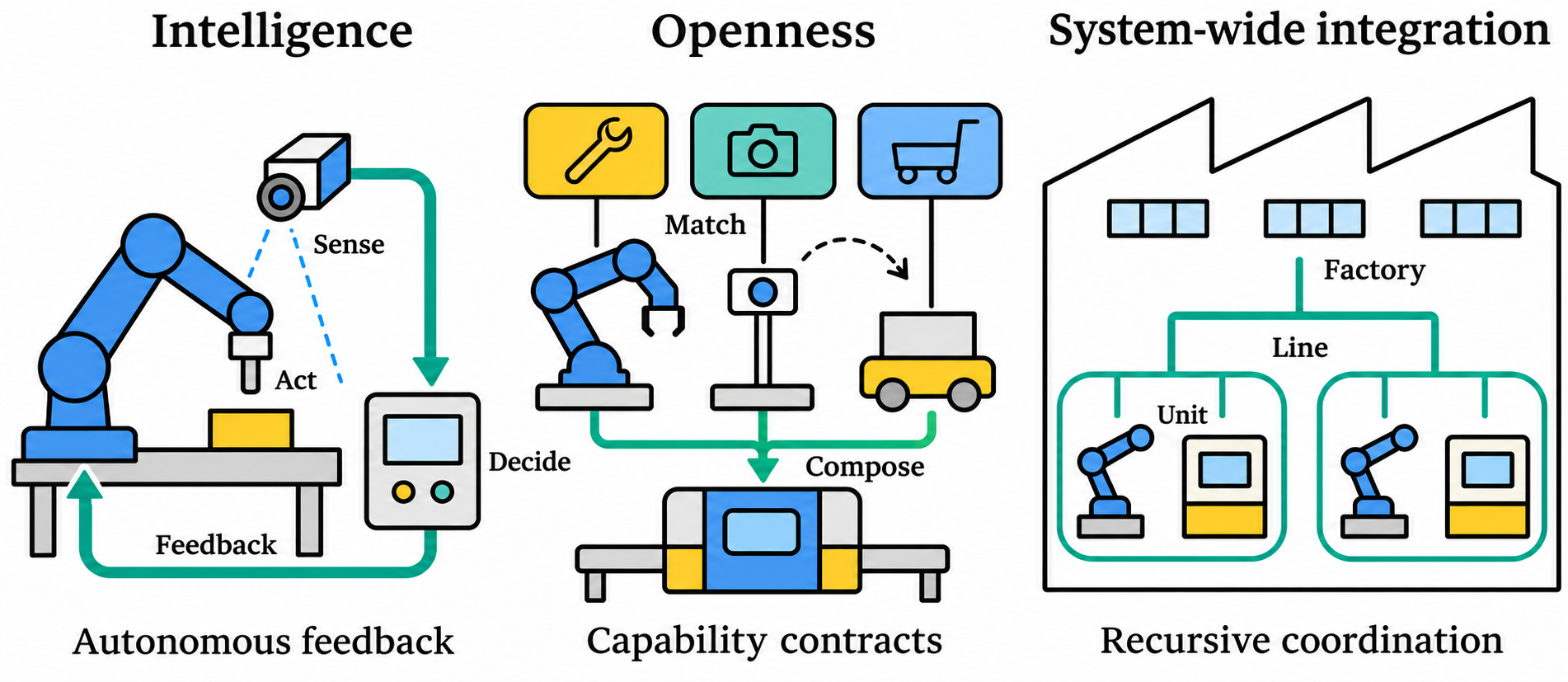}
    \caption{The three panels show the primary contributions of autonomous feedback, capability contracts, and recursive coordination to intelligence, openness, and system-wide integration, respectively.}
    \label{fig:raa-manufacturing}
\end{figure*}

\section{Advancing Intelligent Manufacturing}\label{sec:manufacturing}

This section discusses the role of the proposed RAA framework in next-generation intelligent production. It places RAA within the evolution of manufacturing paradigms and explains how its mechanisms support intelligence, openness, and system-wide integration.

\subsection{Manufacturing Paradigms}

The design addresses a broader development in manufacturing. Early mechanization and electrification increased the physical capacity of equipment. Computer and numerical-control technologies later made it possible to encode and repeat production operations. Digital tools also expanded into design and management. Computer-aided design (CAD) represented product information. Enterprise resource planning (ERP) supported the planning of orders, materials, and resources. Manufacturing execution systems (MES) organized and tracked shop-floor work~\cite{zhou2019hcps}. These developments improved individual activities, even when information remained separated between systems.

Networks extended information exchange across equipment, departments, and enterprises. The Industrial Internet of Things, cyber-physical systems, and digital twins further connected field conditions to planning and control~\cite{soori2023,zhou2018ngim}. Such connections make coordinated work possible, but they do not remove the need for a decision process when circumstances fall outside a fixed plan.

Zhou et al.~\cite{zhou2018ngim} and Zang et al.~\cite{zang2018paradigms} describe three basic paradigms: digital manufacturing, digital and networked manufacturing, and new-generation intelligent manufacturing. The first embeds existing knowledge in software and programs. The second connects information and resources. The third integrates new-generation AI with manufacturing so that systems can learn, apply knowledge, and adapt decisions to operating conditions.

The first two paradigms leave two related difficulties. Separate digital systems may use incompatible formats or interfaces. People then transfer information and translate plans between them. Connected systems can reduce this separation, but a revised management plan may still fail to produce a timely change in equipment behavior~\cite{wang2018hcps,miit2021smartmanufacturing}. A system may detect a failed machine without being able to reorganize the affected work.

Recent work on large language model (LLM) agents illustrates ways to strengthen the link from production requirements to action. Xia et al.~\cite{xia2023llm} connect task descriptions to planning and equipment skill invocation in a modular production setting. Fan et al.~\cite{fan2025} examine LLM support for autonomous industrial robotics. These developments motivate a framework that connects local decision capability to the organization of multiple production units.

\subsection{Alignment with Manufacturing Requirements}

Intelligence, openness, and system-wide integration are key characteristics in Zhou's account of next-generation intelligent manufacturing~\cite{zhou2026ngim}. Figure~\ref{fig:raa-manufacturing} illustrates how the RAA
framework supports intelligence, openness, and system-wide
integration through its engineering mechanisms.

\paragraph{Intelligence}
A production unit must interpret its objective, act using current conditions, and check whether the result is acceptable. When it detects a deviation, it adjusts the plan within its authority. Experience can improve later decisions through controlled updates. RAA connects these functions in the local agent. This is intended to reduce dependence on fixed programs and repeated manual intervention when conditions change~\cite{wang2018hcps}.

\paragraph{Openness}
An open production organization needs to know what a new unit can provide and under what conditions. Standardized descriptions, including the Asset Administration Shell, support a common representation of equipment information~\cite{aas2023}. RAA uses capability contracts to connect such descriptions to task selection and execution requirements. Material, time, and resource needs must be understood before capabilities can be combined.

\paragraph{System Integration}
Production units need to contribute to the overall delivery objective. They must also cooperate with activities such as design, logistics, and services~\cite{zhou2018ngim,miit2021smartmanufacturing}. A station cannot plan only for its own convenience when another task depends on its output. Shared resources and handoffs connect local choices to broader results.
RAA carries goals downward and brings results upward through the same pattern at each level. A team-like organization resolves issues within its scope and escalates those that affect a broader group. This mirrors the functional links between operators, teams, workshops, and factories. Coordination gives local action an organizational context and helps prevent local optimization from undermining the overall task.

\subsection{Deployment Scope}

These three requirements reinforce one another. A unit that can adapt but cannot describe its capabilities is difficult to compose into larger tasks. Connected units that cannot decide locally still depend on external instructions. Autonomous units without coordination can interfere with one another. Local feedback, capability contracts, and recursive organization are intended to address these needs together.

Deployment can start with a workstation or production-line segment that offers clear practical value. The enterprise can establish the sensing, connectivity, and decision capability needed for that task, then extend the organization using validated units. Factory-wide transformation need not be a prerequisite for every local application. This is relevant to enterprises with limited funding or technical staff~\cite{oecd2021sme}.

\section{Conclusion}\label{sec:conclusion}

Production intelligence concerns how a system selects actions under changing conditions and revises them through feedback. A CAS perspective brings two mechanisms into focus. Individual autonomy forms local decisions using available information. Collective coordination manages the dependencies that determine whether those decisions can work together. Learning and adaptation improve both mechanisms over time.
The RAA framework turns this account into an organization of agents and coordinated groups. Goal contracts guide local work. Capability contracts support participant selection and composition. Recursive interfaces connect equipment, production lines, and factories, while feedback connects results to decisions at each level. Checked execution, local recovery, escalation, and controlled learning preserve the role of operating boundaries throughout this design.
RAA provides an engineering approach to next-generation intelligent manufacturing. It connects local autonomy with recursive coordination to support intelligence, openness, and system-wide integration.

\begingroup
\small
\raggedright
\bibliographystyle{plain}
\bibliography{references}
\endgroup
\end{document}